\documentclass[final]{cvpr}

\usepackage{times}
\usepackage{epsfig}
\usepackage{graphicx}
\usepackage{subfigure}
\usepackage{capt-of}

\usepackage{amsmath,amssymb,amsfonts,dsfont,bm,mathrsfs}
\usepackage{booktabs,multirow,adjustbox}
\usepackage[pagebackref=true,breaklinks=true,colorlinks,bookmarks=false]{hyperref}
\usepackage[table]{xcolor}
\usepackage{collcell}
\usepackage{xifthen}
\makeatletter
\@namedef{ver@everyshi.sty}{}
\makeatother
\usepackage{pgf}
\def\ColorMode{hsb}
\newcommand{\ColCell}[1]{
  \ifthenelse{\isempty{#1}}{}{
    \pgfmathparse{#1<2?1:0}                          
      \ifnum\pgfmathresult=0\relax\color{white}\fi
    \pgfmathsetmacro\compA{0}                        
    \pgfmathsetmacro\compB{(abs(#1)==1?0:abs(#1))/1} 
    \pgfmathsetmacro\compC{1}                        
    \edef\x{\noexpand\centering\noexpand\cellcolor[\ColorMode]{\compA,\compB,\compC}}\x #1
  }
}
\newcolumntype{C}[1]{>{\collectcell\ColCell}m{#1}<{\endcollectcell}}  

\newcommand{\NumAttributes}{5}

\newcommand{\R}{\mathbb{R}}       
\newcommand{\0}{{\rm\bf 0}}       
\newcommand{\z}{{\rm\bf z}}       
\newcommand{\Z}{\mathcal{Z}}      
\newcommand{\img}{{\rm\bf I}}     
\newcommand{\Img}{\mathcal{I}}    
\newcommand{\A}{{\rm\bf A}}       
\renewcommand{\b}{{\rm\bf b}}     
\newcommand{\y}{{\rm\bf y}}       
\newcommand{\n}{{\rm\bf n}}       
\newcommand{\N}{{\rm\bf N}}       
\newcommand{\argmax}{\mathop{\arg\max}}  

\begin{document}

\title{Closed-Form Factorization of Latent Semantics in GANs}

\author{
Yujun Shen \qquad Bolei Zhou\\
The Chinese University of Hong Kong\\
{\tt\small \{sy116, bzhou\}@ie.cuhk.edu.hk}
}

\twocolumn[{
\renewcommand\twocolumn[1][]{#1}
\maketitle
\begin{center}
  \includegraphics[width=1.0\linewidth]{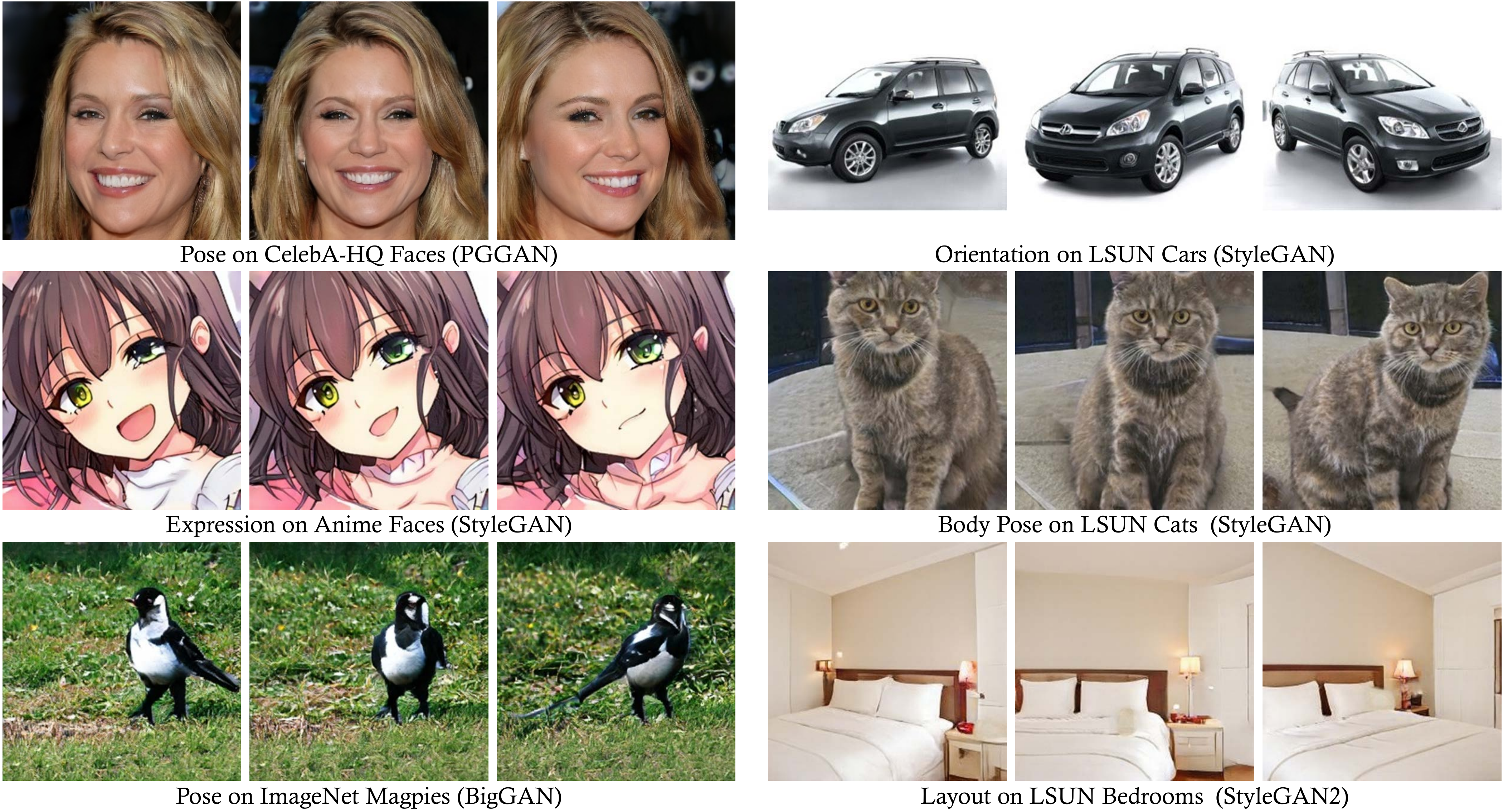}
  \vspace{-12pt}
  \captionof{figure}{
    \textbf{Versatile interpretable directions of the latent space unsupervisedly discovered in different GAN models} including PGGAN~\cite{pggan}, StyleGAN~\cite{stylegan}, BigGAN~\cite{biggan}, and StyleGAN2~\cite{stylegan2}.
    For each set of images, the middle one is the original output, while the left and the right are the output images by moving the latent code toward and backward the interpretable direction found by SeFa.
  }
  \label{fig:teaser}
  \vspace{5pt}
\end{center}
}]

\begin{abstract}
A rich set of interpretable dimensions has been shown to emerge in the latent space of the Generative Adversarial Networks (GANs) trained for synthesizing images.
In order to identify such latent dimensions for image editing, previous methods typically annotate a collection of synthesized samples and train linear classifiers in the latent space.
However, they require a clear definition of the target attribute as well as the corresponding manual annotations, limiting their applications in practice.
In this work, we examine the internal representation learned by GANs to reveal the underlying variation factors in an unsupervised manner.
In particular, we take a closer look into the generation mechanism of GANs and further propose a closed-form factorization algorithm for latent semantic discovery by directly decomposing the pre-trained weights.
With a lightning-fast implementation, our approach is capable of not only finding semantically meaningful dimensions comparably to the state-of-the-art supervised methods, but also resulting in far more versatile concepts across multiple GAN models trained on a wide range of datasets.%
\footnote{Project page is at \url{https://genforce.github.io/sefa/}.}
\end{abstract}

\section{Introduction}\label{sec:introcution}
Generative Adversarial Networks (GANs)~\cite{gan} have achieved tremendous success in image synthesis~\cite{pggan,stylegan,biggan,stylegan2}.
It has been recently found that when learning to synthesize images, GANs spontaneously represent multiple interpretable attributes in the latent space~\cite{goetschalckx2019ganalyze,gansteerability,interfacegan,plumerault2020controlling,yang2019semantic}, such as gender for face synthesis~\cite{interfacegan} and lighting condition for scene synthesis~\cite{yang2019semantic}.
By properly identifying these semantics, we can reuse the knowledge learned by GANs to reasonably control the image generation process, enabling a wide range of editing applications, like face manipulation~\cite{interfacegan2,gu2020image} and scene editing~\cite{yang2019semantic,zhu2020indomain}.

The crux of interpreting the latent space of GANs is to find the meaningful directions in the latent space corresponding to the human-understandable concepts~\cite{goetschalckx2019ganalyze,gansteerability,interfacegan,plumerault2020controlling,yang2019semantic}.
Through that, moving the latent code towards the identified direction can accordingly change the semantic occurring in the output image.
However, due to the high dimensionality of the latent space as well as the large diversity of image semantics, finding valid directions in the latent space is extremely challenging.

Existing supervised approaches typically first randomly sample a large amount of latent codes, then synthesize a collection of images and annotate them with some pre-defined labels, and finally use these labeled samples to learn a classifier in the latent space.
To get the labels for training, they either employ pre-trained attribute predictors~\cite{goetschalckx2019ganalyze,interfacegan,yang2019semantic} or utilize some simple statistical information of the image (\textit{e.g.}, object position and color tone)~\cite{gansteerability,plumerault2020controlling}.
Several limitations rise from the above supervised training process.
Firstly, relying on pre-defined classifiers hinders the algorithm from being applied to the case where the classifiers are not available or difficult to train.
On the other hand, sampling is both time-consuming and unstable, \textit{e.g.}, a different collection of synthesized data may lead to a different training result.
Some very recent studies explore the unsupervised discovery of interpretable GAN semantics~\cite{voynov2020unsupervised,harkonen2020ganspace}, but they also require model training~\cite{voynov2020unsupervised} or data sampling~\cite{harkonen2020ganspace}.

In this work, we propose a novel algorithm to discover the latent semantic directions learned by GANs, which is \textit{independent of any kind of training or sampling}.
We call it \textit{SeFa} as the short for \textit{Semantic Factorization}.
Instead of relying on the synthesized samples as an intermediate step, SeFa takes a deep look into the generation mechanism of GANs to examine the relation between the image variation and the internal representation.
In fact, GANs project a latent code to a photo-realistic image step by step (or say layer by layer), where each step learns a projection from one space to another.
Many explanatory factors originate in such process.
Thus we investigate the first projection step that directly acts on the latent space we want to study.
We propose a \textit{closed-form} method that can identify versatile semantics from the latent space by \textit{merely using the pre-trained weights of the generator}.
More importantly, these variation factors, unsupervisedly found by SeFa, are accurate and in a wider range compared to the state-of-the-art supervised approaches.
We demonstrate some interesting manipulation results using the discovered semantics in Fig.~\ref{fig:teaser}.
For instance, we can rotate the object in an image without knowing its underlying 3D model or pose label.
Extensive experiments suggest that our approach is efficient and applicable to most popular GAN models (\textit{e.g.}, PGGAN~\cite{pggan}, StyleGAN~\cite{stylegan}, BigGAN~\cite{biggan}, and StyleGAN2~\cite{stylegan2}) that are trained on different datasets.

\subsection{Related Work}\label{subsec:related-work}

\noindent\textbf{Generative Adversarial Networks.}
GAN~\cite{gan} has significantly advanced image synthesis in recent years~\cite{dcgan,wgan,pggan,biggan,stylegan,stylegan2}.
The generator in GANs can take a randomly sampled latent code as the input and output a high-fidelity image through adversarial learning.
Existing GAN models are commonly built on deep convolutional neural networks where the latent code is fed into the first convolution layer using an affine transformation~\cite{dcgan,wgan,pggan}.
Recently, this idea is improved by the style-based generator~\cite{stylegan,stylegan2} where the latent code is mapped to layer-wise style codes and then fed into each convolution layer through Adaptive Instance Normalization (AdaIN)~\cite{adain} operation.

\noindent\textbf{Latent Semantic Interpretation.}
Generative models show great potential in learning variation factors from observed data.
Chen \textit{et al.}~\cite{infogan} and Higgins \textit{et al.}~\cite{betavae} propose to add regularizers into the training process to explicitly learn an interpretable factorized representation.
Recent work has found that the native GANs, without any constraints or regularizers, are able to automatically encode various semantics in the intermediate feature space~\cite{bau2019gandissect} and the initial latent space~\cite{goetschalckx2019ganalyze,gansteerability,interfacegan,yang2019semantic}.
However, these methods are usually performed in a supervised fashion, which requires sampling a collection of images and labeling them to train a classifier.
Thus they heavily rely on the attribute predictors or human annotators to get the label.
Some concurrent work studies unsupervised semantic discovery in GANs.
Voynov and Babenko~\cite{voynov2020unsupervised} jointly learn a candidate matrix and a classifier such that the semantic directions in the matrix can be properly recognized by the classifier.
H\"{a}rk\"{o}nen \textit{et al.}~\cite{harkonen2020ganspace} perform PCA on the sampled data to find primary directions in the latent space.
However, they still require model training~\cite{voynov2020unsupervised} and data sampling~\cite{harkonen2020ganspace}.
Differently, we study the generation mechanism of GANs and propose a \textit{closed-form} factorization method, which is independent of any kind of training or sampling.

\section{Method}\label{sec:method}
We introduce SeFa, a closed-form method to discover latent interpretable directions in GANs.
By taking a close look into the generation mechanism of GANs, SeFa can identify semantically meaningful directions in the latent space efficiently by decomposing the model weights.

\subsection{Preliminaries}\label{subsec:prelimiaries}

\noindent\textbf{Generation Mechanism of GANs.}
The generator $G(\cdot)$ in GANs learns the mapping from the $d$-dimensional latent space $\Z\subseteq\R^d$ to a higher dimensional image space $\Img\subseteq\R^{H \times W \times C}$, as $\img=G(\z)$.
Here, $\z\in\Z$ and $\img\in\Img$ denote the input latent code and the output image respectively.
State-of-the-art GAN models~\cite{dcgan,pggan,biggan,stylegan,stylegan2} typically adopt convolutional neural networks as the generator architecture.
Consisting of multiple layers, $G(\cdot)$ projects the starting latent space to the final image space step by step.
Each step learns a transformation from one space to another.
We focus on examining the first step, which directly acts on the latent space we would like to explore.
In particular, it can be formulated as an affine transformation, like most GANs~\cite{dcgan,pggan,biggan,stylegan,stylegan2} have done, as
\begin{align}
  G_1(\z) \triangleq \y = \A\z + \b, \label{eq:affine}
\end{align}
where $\y\in\R^m$ is the $m$-dimensional projected code.
$\A\in\R^{m \times d}$ and $\b\in\R^m$ denote the weight and bias used in the first transformation step $G_1(\cdot)$ respectively.

\vspace{2pt}
\noindent\textbf{Manipulation Model in GAN Latent Space.}
The latent space of GANs has recently been shown to encode rich semantic knowledge~\cite{goetschalckx2019ganalyze,gansteerability,interfacegan,yang2019semantic}.
These semantics can be further applied to image editing with the vector arithmetic property~\cite{dcgan}.
More concretely, prior work~\cite{goetschalckx2019ganalyze,interfacegan,yang2019semantic,voynov2020unsupervised,harkonen2020ganspace} proposed to use a certain direction $\n\in\R^d$ in the latent space to represent a semantic concept.
After identifying a semantically meaningful direction, the manipulation can be achieved via the following model
\begin{align}
  \mathtt{edit}(G(\z)) = G(\z') = G(\z + \alpha\n), \label{eq:manipulation}
\end{align}
which is commonly used in the existing approaches~\cite{goetschalckx2019ganalyze,interfacegan,yang2019semantic,voynov2020unsupervised,harkonen2020ganspace}.
Here, $\mathtt{edit}(\cdot)$ denotes the editing operation.
In other words, we can alter the target semantic by linearly moving the latent code $\z$ along the identified direction $\n$.
$\alpha$ indicates the manipulation intensity.

\subsection{Unsupervised Semantic Factorization}\label{subsec:semantic-discovery}
Our goal is to reveal the explanatory factors (\textit{i.e.}, the direction $\n$ in Eq.~\eqref{eq:manipulation}) from the latent space of GANs.
As discussed above, the generator in GANs can be viewed as a multi-step function that gradually projects the latent space to the image space.
Let us take a closer look into the first projection step, as suggested in Eq.~\eqref{eq:affine}.
Under its formulation of affine transformation, the manipulation model in Eq.~\eqref{eq:manipulation} can be simplified as
\begin{align}
  \y' \triangleq G_1(\z') &= G_1(\z + \alpha\n) \nonumber \\
                          &= \A\z + \b + \alpha\A\n = \y + \alpha\A\n. \label{eq:new-manipulation}
\end{align}

We observe from Eq.~\eqref{eq:new-manipulation} that the manipulation process is instance independent.
In other words, given any latent code $\z$ together with a certain latent direction $\n$, the editing can be always achieved by adding the term $\alpha\A\n$ onto the projected code after the first step.
From this perspective, the weight parameter $\A$ should contain the essential knowledge of the image variation.
Thus we aim to discover important latent directions by decomposing $\A$.

To this end, we propose an \textit{unsupervised} approach, which is \textit{independent of data sampling and model training}, for semantic factorization by solving the following optimization problem
\begin{align}
  \n^* = \argmax_{\{\n\in\R^d:\ \n^T\n = 1\}} ||\A\n||_2^2,  \label{eq:single-optimization}
\end{align}
where $||\cdot||_2$ denotes the $l_2$ norm.
This problem aims at finding the directions that can cause large variations after the projection of $\A$.
Intuitively, if some direction $\n'$ is projected to a zero-norm vector, \textit{i.e.}, $\A\n'=\0$, the editing operation in Eq.~\eqref{eq:new-manipulation} turns into $\y' = \y$, which will keep the output synthesis unchanged, let clone alter the semantics occurring in it.

When the case comes to finding $k$ most important directions $\{\n_1, \n_2, \cdots, \n_k\}$, we expand Eq.~\eqref{eq:single-optimization} into
\begin{align}
  \N^* = \argmax_{\{\N\in\R^{d\times k}:\ \n_i^T\n_i = 1\ \forall i=1,\cdots,k\}} \sum_{i=1}^k ||\A\n_i||_2^2,  \label{eq:optimization}
\end{align}
where $\N = [\n_1, \n_2, \cdots, \n_k]$ correspond to the top-$k$ semantics.
To solve this problem, we introduce the Lagrange multipliers $\{\lambda_i\}_{i=1}^k$ into Eq.~\eqref{eq:optimization} as
\begin{align}
  \N^* &= \argmax_{\N\in\R^{d\times k}} \sum_{i=1}^k ||\A\n_i||_2^2 - \sum_{i=1}^k\lambda_i(\n_i^T\n_i - 1) \nonumber \\
       &= \argmax_{\N\in\R^{d\times k}} \sum_{i=1}^k (\n_i^T\A^T\A\n_i - \lambda_i\n_i^T\n_i + \lambda_i). \label{eq:lagrange}
\end{align}
By taking the partial derivative on each $\n_i$, we have
\begin{align}
  2\A^T\A\n_i - 2\lambda_i\n_i = 0. \label{eq:solution}
\end{align}

All possible solutions to Eq.~\eqref{eq:solution} should be the eigenvectors of the matrix $\A^T\A$.
To get the maximum objective value and make $\{\n_i\}_{i=1}^k$ distinguishable from each other, we choose columns of $\N$ as the eigenvectors of $\A^T\A$ associated with the $k$ largest eigenvalues.

\begin{figure*}[t]
  \centering
  \includegraphics[width=1.0\linewidth]{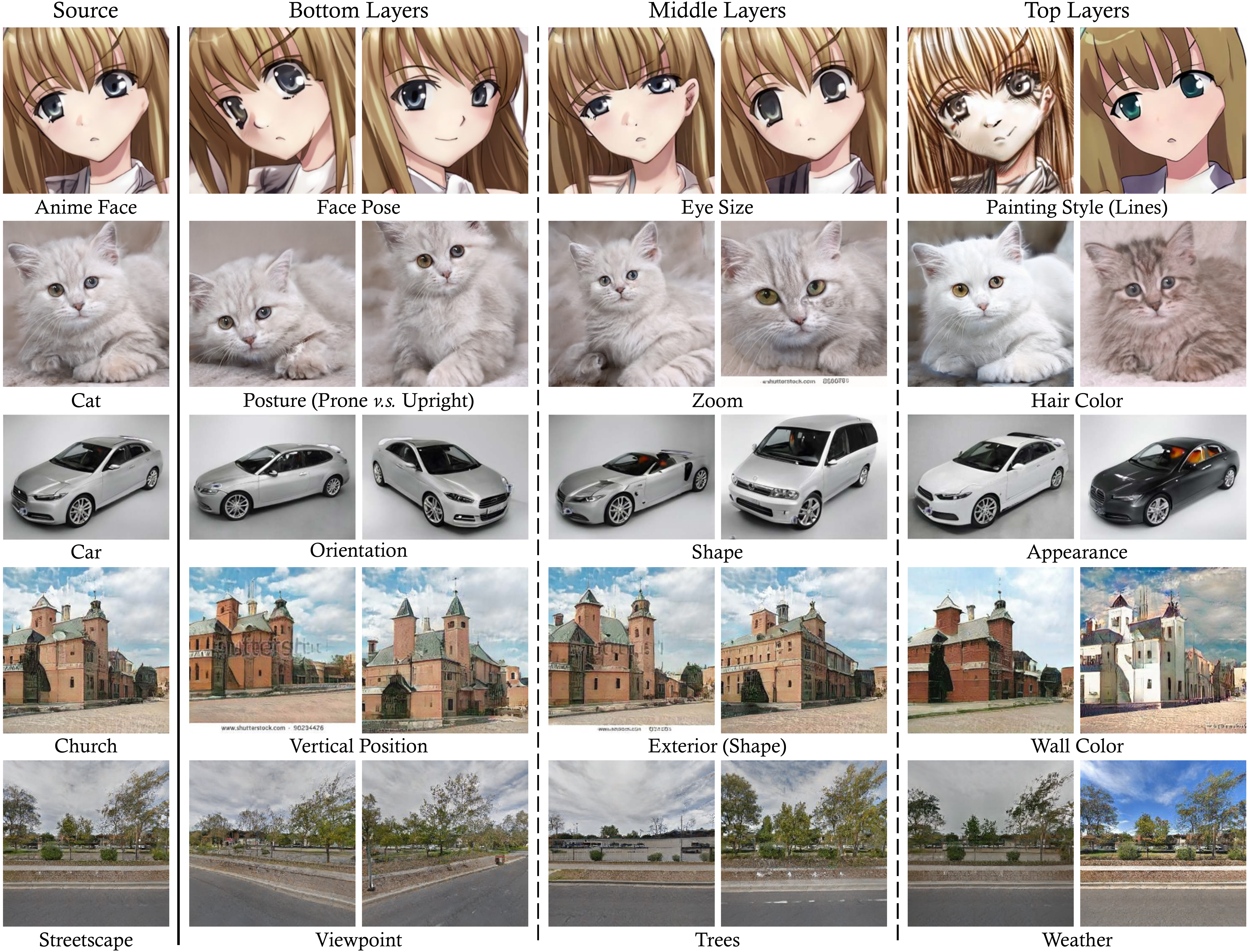}
  \caption{
   \textbf{Hierarchical interpretable directions} discovered in the style-based generators, \textit{i.e.}, StyleGAN~\cite{stylegan} and StyleGAN2~\cite{stylegan2}.
    Among them, the streetscapes model is trained with StyleGAN2, while the others are using StyleGAN.
  }
  \label{fig:layerwise}
  \vspace{0pt}
\end{figure*}

\begin{figure*}[t]
  \centering
  \includegraphics[width=1.0\linewidth]{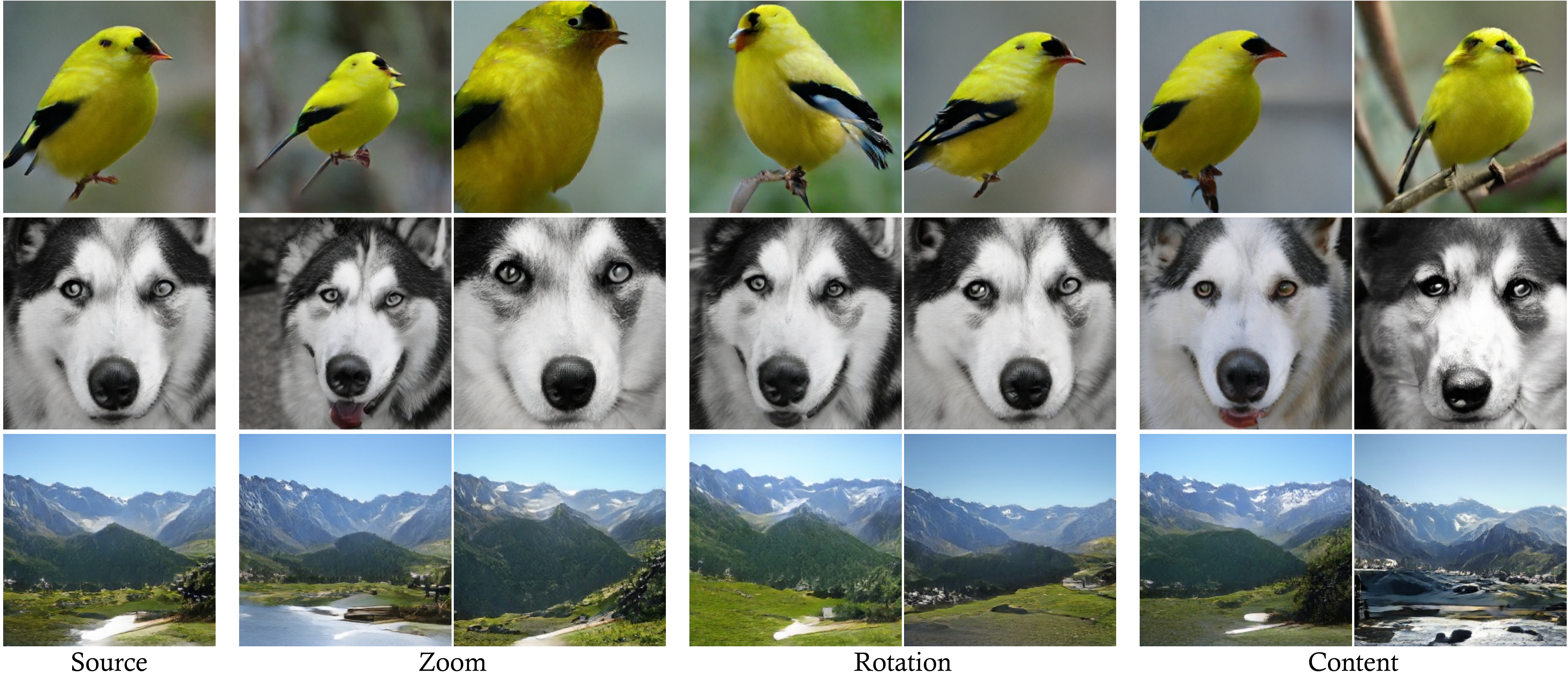}
  \caption{
    \textbf{Diverse interpretable directions} found in the BigGAN~\cite{biggan}, which is conditionally trained on ImageNet~\cite{imagenet}.
    These semantics are further used to manipulate images from different categories.
  }
  \label{fig:biggan}
  \vspace{-5pt}
\end{figure*}

\subsection{Implementation on GAN Models}\label{subsec:implementation}
In Sec.~\ref{subsec:semantic-discovery}, we propose a closed-form algorithm, termed as SeFa, to factorize the latent semantics learned by GANs.
Our algorithm can be performed \textit{in a completely unsupervised fashion by efficiently investigating the weights of a pre-trained GAN generator}.
In this part, we introduce how our approach is applied to the state-of-the-art GAN models, such as PGGAN~\cite{pggan}, StyleGAN~\cite{stylegan}, and BigGAN~\cite{biggan}.

\vspace{2pt}
\noindent\textbf{PGGAN.}
PGGAN~\cite{pggan} is a representative of the conventional generator, where the input latent code is firstly mapped into a spatial feature map and then projected to an image with a sequence of convolution layers.
For this kind of generator structure, SeFa studies the transformation from the latent code to the feature map.

\vspace{2pt}
\noindent\textbf{StyleGAN.}
StyleGAN~\cite{stylegan} proposes the style-based generator, which feeds the latent code into each convolution layer.
In particular, for each layer, the latent code is transformed to a style code, which is used to alter the channel-wise mean and variance of the feature map through Adaptive Instance Normalization (AdaIN)~\cite{adain}.
For this GAN type, we investigate the transformation from the latent code to the style code.
Note that our algorithm is flexible such that it supports interpreting all or any subset of layers.
For this purpose, we concatenate the weight parameters (\textit{i.e.}, $\A$ in Eq.~\eqref{eq:affine}) from all target layers along the first axis, forming a larger transformation matrix.

\vspace{2pt}
\noindent\textbf{BigGAN.}
BigGAN~\cite{biggan} is a large-scale GAN model primarily designed for conditional generation.
The latent code is both mapped to the initial feature map and fed into each convolution layer.
Hence, the analysis on BigGAN can be viewed as a combination of the above two types of GANs.

\section{Experiments}\label{sec:experiments}
We evaluate our closed-form algorithm on a wide range of models to discover interpretable directions.
We also compare SeFa with existing supervised and unsupervised alternatives to demonstrate its effectiveness.

\subsection{Results on Diverse Models and Datasets}\label{subsec:generalization}
We conduct experiments on the state-of-the-art GAN models, such as StyleGAN~\cite{stylegan}, BigGAN~\cite{biggan}, and StyleGAN2~\cite{stylegan2}.
They are trained on different datasets, including human faces (FF-HQ~\cite{stylegan}), anime faces~\cite{animeface}, scenes and objects (LSUN~\cite{lsun}), streetscapes~\cite{streetscapes}, and ImageNet~\cite{imagenet}.%
\footnote{We have collected a model zoo consisting of various types of GANs. SeFa can be easily applied to interpreting these models benefiting from its efficient implementation (\textit{i.e.}, less than 1 second for one model). Please refer to the \href{https://www.youtube.com/watch?v=OFHW2WbXXIQ}{demo video} for diverse and continuous manipulation results.}

\begin{figure}[t]
  \centering
  \includegraphics[width=1.0\linewidth]{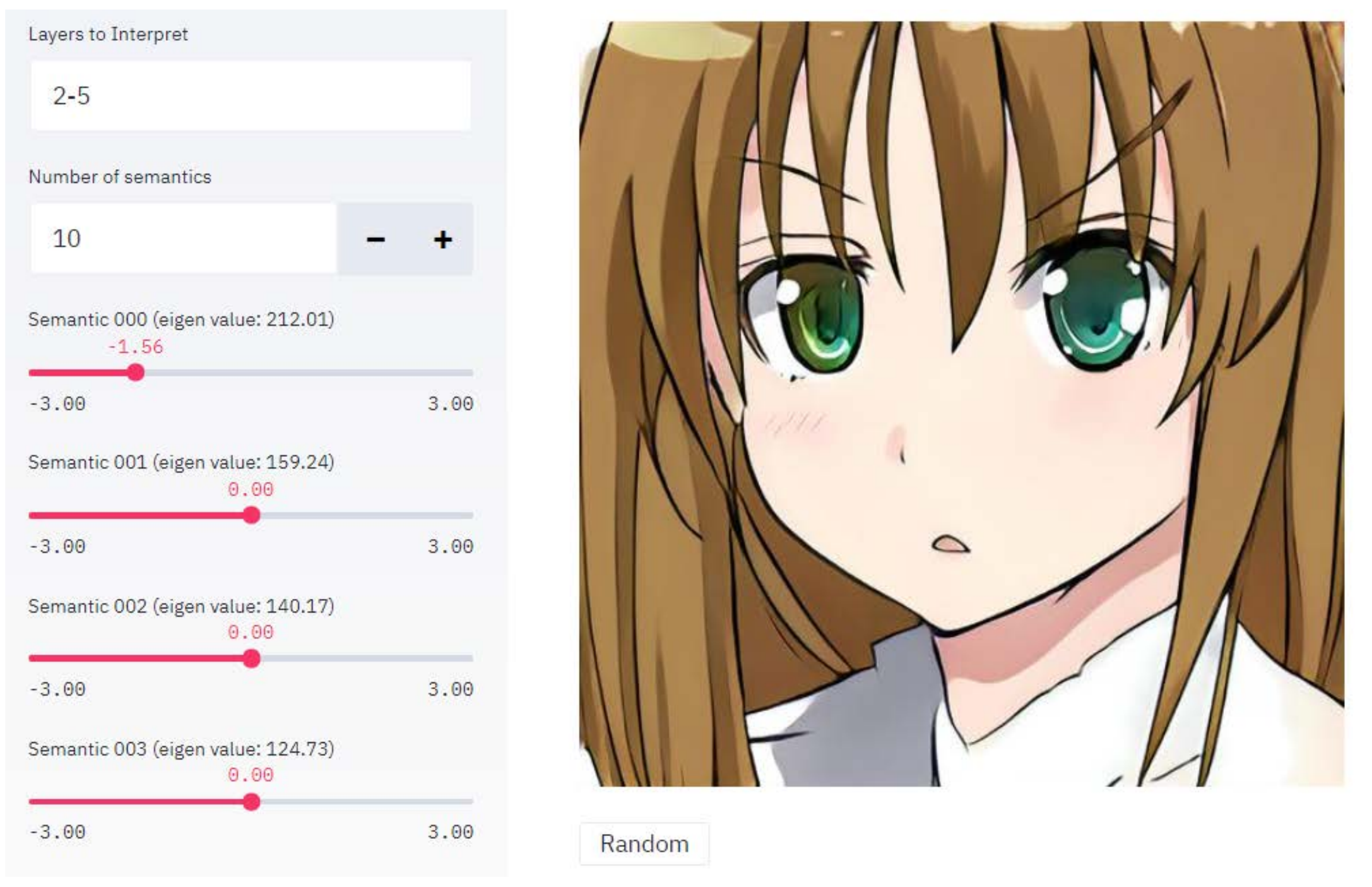}
  \caption{
    \textbf{Interface} for interactive editing.
  }
  \label{fig:interface}
  \vspace{-2pt}
\end{figure}

\vspace{2pt}
\noindent\textbf{Interactive Editing by Tuning Interpretable Directions.}
Our algorithm is performed in a completely unsupervised manner, hence we do not rely on any auxiliary predictors.
After discovering the important directions by decomposing the model weights, we can interact with the GAN model for collaborative content editing.
Thus we develop an interface to facilitate human-model interaction, as shown in Fig.~\ref{fig:interface}.
Meanwhile, with the help of this interface, users can easily annotate the identified semantics.

\vspace{2pt}
\noindent\textbf{Results on StyleGAN.}
As described in Sec.~\ref{subsec:implementation}, our algorithm can interpret a subset of layers in the style-based generators~\cite{stylegan,stylegan2}.
We evaluate SeFa on the models trained on a wide range of datasets, including anime faces, objects, scenes, and streetscapes.
In particular, we interpret a target model at the levels of bottom layers, middle layers, and top layers respectively.
Fig.~\ref{fig:layerwise} shows the versatile semantic directions found in these models.
We noticeably find that they are organized as a hierarchy, which is consistent with the observations from prior work~\cite{stylegan,yang2019semantic}.
Taking cars as an example, bottom layers tend to control the rotation, middle layers determine the shape, while top layers correspond to the color.
We further conduct a user study to see how the variation factors found by SeFa align with human perception.
Here, questions are asked to 10 annotators.
As suggested in Tab.~\ref{tab:user-study}, SeFa can indeed find human-understandable concepts, even from some particular layers in GAN models.

\setlength{\tabcolsep}{5pt}
\begin{table}[t]
  \caption{
    User study.
    We randomly generate $2K$ images for each dataset, and use the Top-50 eigen directions from each level of layers to manipulate these images.
    Numbers in brackets indicate the index of the layers to interpret.
    Users are asked how many directions result in \textit{obvious} content change (numerator) and how many directions are semantically meaningful (denominator).
  }
  \vspace{2pt}
  \label{tab:user-study}
  \centering\small
  \begin{tabular}{lccc}
    \toprule
    Dataset                         & Bottom (0-1) & Middle (2-5) & Top (6-) \\ \midrule
    Anime Face~\cite{animeface}     &        12/12 &        26/26 &    38/50 \\
    LSUN Cat~\cite{lsun}            &        14/15 &        21/28 &    47/50 \\
    LSUN Car~\cite{lsun}            &        10/10 &        16/22 &    22/34 \\
    LSUN Church~\cite{lsun}         &        15/15 &        18/26 &    48/50 \\
    Streetscape~\cite{streetscapes} &         9/9  &        12/18 &    15/36 \\ \bottomrule
  \end{tabular}
  \vspace{-10pt}
\end{table}

\begin{figure*}[t]
  \centering
  \includegraphics[width=1.0\linewidth]{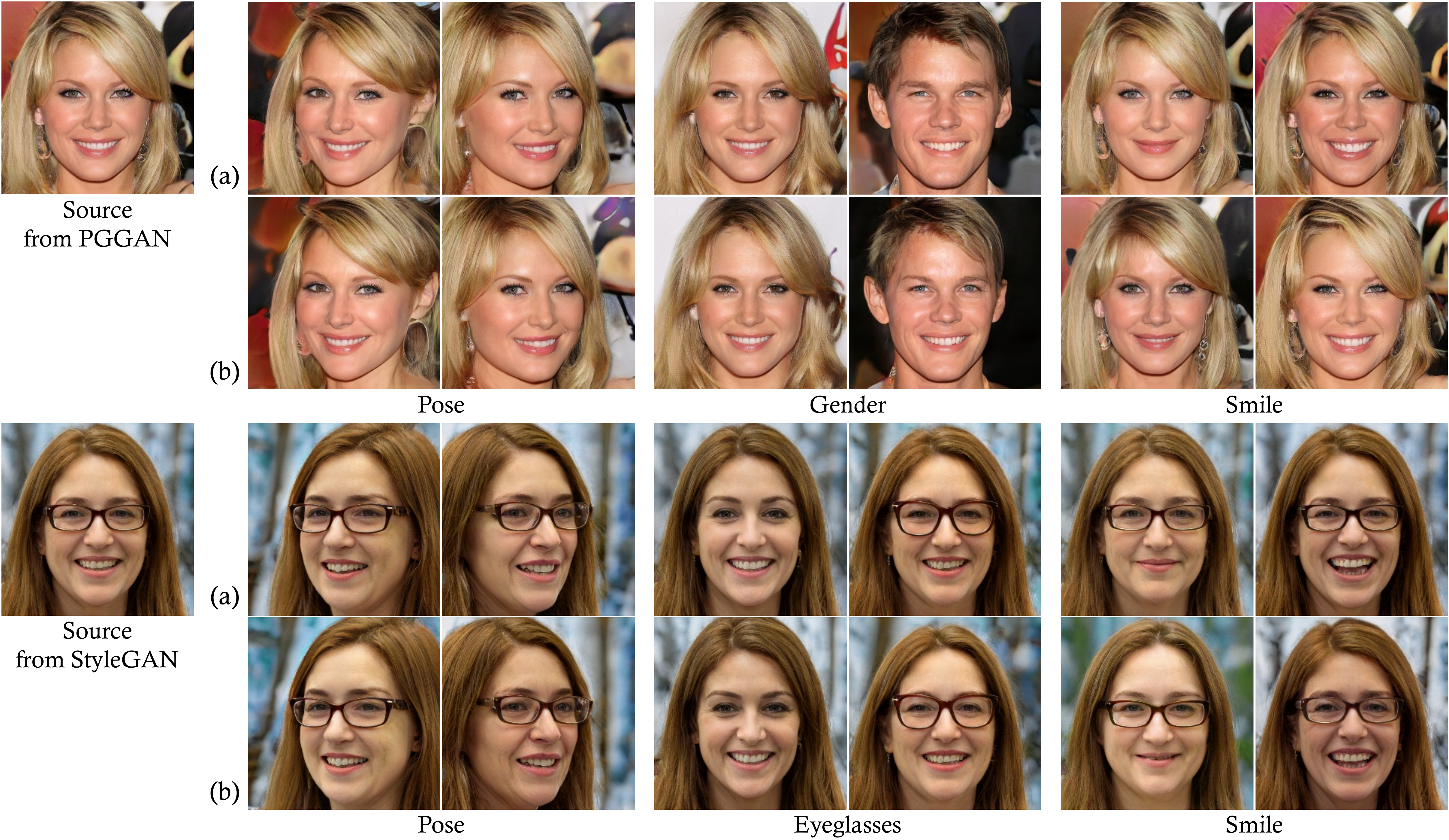}
  \caption{
    Qualitative comparison of the latent semantics found by (a) the supervised method, InterFaceGAN~\cite{interfacegan} and (b) our \textit{closed-form} solution, SeFa, where SeFa achieves similar performance to InterFaceGAN.
    PGGAN trained on CelebA-HQ~\cite{pggan} and StyleGAN trained on FF-HQ~\cite{stylegan} are used as the target models to interpret.
  }
  \label{fig:comparison-interfacegan}
  \vspace{0pt}
\end{figure*}

\setlength{\tabcolsep}{1pt}
\begin{table*}[t]
  \caption{
    \textbf{Re-scoring analysis} of the semantics identified by InterFaceGAN~\cite{interfacegan} and SeFa from the PGGAN model trained on CelebA-HQ dataset~\cite{pggan}.
    Each row evaluates how the semantic scores change after moving the latent code along a certain direction.
  }
  \label{tab:rescoring}
  \vspace{2pt}
  \centering
  \subfigure[InterFaceGAN~\cite{interfacegan}, which is supervised.]{
    \centering\small
    \begin{tabular}{c*{\NumAttributes}{C{28pt}}}
      \multicolumn{1}{c}{} &
      \multicolumn{1}{c}{Pose} &
      \multicolumn{1}{c}{Gender} &
      \multicolumn{1}{c}{Age} &
      \multicolumn{1}{c}{Glasses} &
      \multicolumn{1}{c}{Smile}                       \\
      Pose    &  0.53 & -0.06 & -0.09 & -0.01 &  0.05 \\
      Gender  & -0.02 &  0.59 &  0.20 &  0.08 & -0.07 \\
      Age     & -0.03 &  0.35 &  0.50 &  0.08 & -0.03 \\
      Glasses & -0.01 &  0.37 &  0.19 &  0.24 &  0.00 \\
      Smile   & -0.01 & -0.07 &  0.03 & -0.01 &  0.60 \\
    \end{tabular}
  }
  \hspace{25pt}
  \subfigure[SeFa, which is unsupervised.]{
    \centering\small
    \begin{tabular}{c*{\NumAttributes}{C{28pt}}}
      \multicolumn{1}{c}{} &
      \multicolumn{1}{c}{Pose} &
      \multicolumn{1}{c}{Gender} &
      \multicolumn{1}{c}{Age} &
      \multicolumn{1}{c}{Glasses} &
      \multicolumn{1}{c}{Smile}                       \\
      Pose    &  0.51 & -0.11 & -0.07 &  0.02 &  0.06 \\
      Gender  &  0.02 &  0.55 &  0.46 &  0.09 & -0.13 \\
      Age     & -0.07 & -0.25 &  0.34 &  0.10 &  0.10 \\
      Glasses &  0.02 &  0.55 &  0.46 &  0.09 & -0.13 \\
      Smile   &  0.03 & -0.03 &  0.15 & -0.16 &  0.42 \\
    \end{tabular}
  }
  \vspace{-10pt}
\end{table*}

\begin{figure*}[t]
  \centering
  \includegraphics[width=1.0\linewidth]{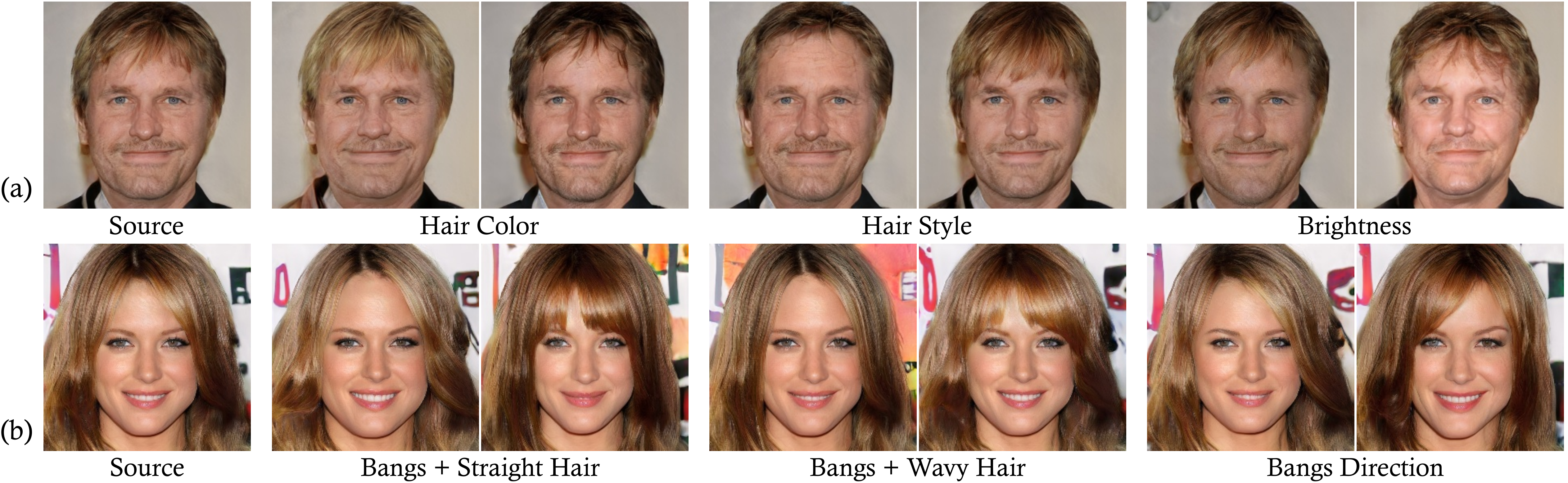}
  \caption{
    (a) Diverse semantics, which can \textit{not} be identified by InterFaceGAN~\cite{interfacegan} due to the lack of semantic predictors.
    (b) Diverse hair styles, which can \textit{not} be described as a binary attribute.
    The PGGAN model trained on CelebA-HQ dataset~\cite{pggan} is used.
  }
  \label{fig:diversity}
  \vspace{-5pt}
\end{figure*}

\begin{figure*}[t]
  \centering
  \includegraphics[width=1.0\linewidth]{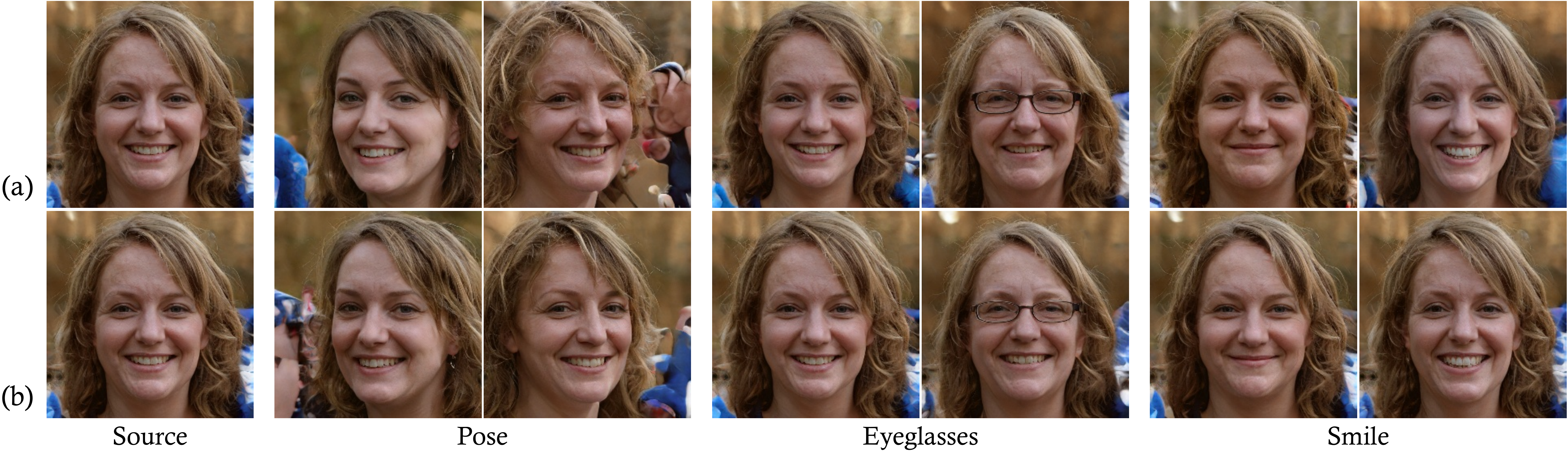}
  \caption{
    Qualitative comparison between (a) GANSpace~\cite{harkonen2020ganspace} and (b) SeFa.
    The StyleGAN model trained on FF-HQ dataset~\cite{stylegan} is used.
  }
  \label{fig:comparison-ganspace}
  \vspace{-5pt}
\end{figure*}

\begin{figure*}[t]
  \centering
  \includegraphics[width=1.0\linewidth]{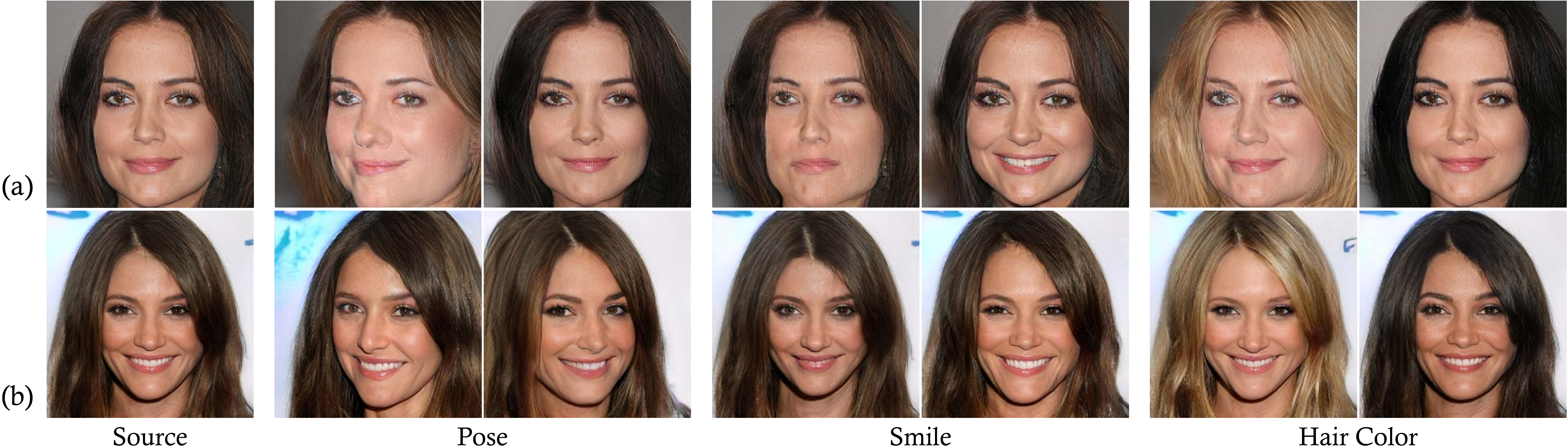}
  \caption{
    Qualitative comparison between (a) Info-PGGAN~\cite{progressive_infogan,infogan} and (b) SeFa.
    The result of the Info-PGGAN model is extracted directly from~\cite{progressive_infogan}, and the official PGGAN model trained on CelebA-HQ dataset~\cite{pggan} is used for SeFa.
  }
  \label{fig:comparison-infogan}
  \vspace{-5pt}
\end{figure*}

\vspace{2pt}
\noindent\textbf{Results on BigGAN.}
We also interpret the large-scale BigGAN~\cite{biggan} model that is conditionally trained on ImageNet~\cite{imagenet}.
BigGAN extends the latent code with a category-derived embedding vector to achieve conditional synthesis.
Here, we only focus on the latent code part for semantic discovery.
Fig.~\ref{fig:biggan} provides some examples.
We can tell that the semantics found by our algorithm can be applied to manipulating images from different categories.
This verifies the generalization ability of SeFa.

\subsection{Comparison with Supervised Approach}\label{subsec:supervise-comparison}
We compare our closed-form algorithm with the state-of-the-art supervised method, InterFaceGAN~\cite{interfacegan}.
We conduct experiments on face synthesis models due to the well definition of facial attributes.
In particular, we make comparison between SeFa and InterFaceGAN on both the conventional generator (\textit{i.e.}, PGGAN~\cite{pggan}) and the style-based generator (\textit{i.e.}, StyleGAN~\cite{stylegan}).

\vspace{2pt}
\noindent\textbf{Qualitative Results.}
Fig.~\ref{fig:comparison-interfacegan} visualizes some manipulation results by using the identified semantics.
We can tell that SeFa achieves similar performance as InterFaceGAN from the perspective of editing pose, gender, eyeglasses, and expression (smile), suggesting its effectiveness.
More importantly, InterFaceGAN requires sampling numerous data and pre-training attribute predictors.
By contrast, SeFa is completely independent of data sampling and model training, which is more efficient and generalizable.

\vspace{2pt}
\noindent\textbf{Re-scoring Analysis.}
For quantitative analysis, we train an attribute predictor on CelebA dataset~\cite{celeba} with ResNet-50 structure~\cite{resnet}, following~\cite{interfacegan}.
With this predictor, we are able to perform re-scoring analysis to quantitatively evaluate whether the identified directions can properly represent the corresponding attributes.
In particular, we randomly sample $2K$ images and manipulate them along a certain discovered direction.
We then use the prepared predictor to check how the semantic score varies in such manipulation process.
Tab.~\ref{tab:rescoring} shows the results where we have three observations.
(i) SeFa can adequately control some attribute, such as pose and gender, similar to InterFaceGAN.
(ii) When altering one semantic, InterFaceGAN shows stronger robustness to other attributes, benefiting from its supervised training manner.
For example, the age and eyeglasses corresponding to the same latent direction identified by SeFa.
That is because the training data is somewhat biased (\textit{i.e.}, older people are more likely to wear eyeglasses), as pointed out by~\cite{interfacegan}.
By contrast, involving labels as the supervision can help learn a more accurate direction to some extent.
(iii) SeFa fails to discover the direction corresponding to eyeglasses.
The reason is that the presence of eyeglasses is not a large variation and hence does not meet the optimization objective in Eq.~\eqref{eq:single-optimization}.

\vspace{2pt}
\noindent\textbf{Diversity Comparison.}
Supervised approach highly depends on the available attribute predictors.
By contrast, our method is more general and can find more diverse semantics in the latent space.
As shown in Fig.~\ref{fig:diversity}~(a), we successfully identify the directions corresponding to hair color, hair style, and brightness.
This surpasses InterFaceGAN since predictors for these attributes are not easy to acquire in practice.
Also, supervised methods are usually limited by the training objective.
For example, InterFaceGAN is proposed to handle binary attributes~\cite{interfacegan}.
In comparison, our method can identify more complex attributes, like the different hair styles shown in Fig.~\ref{fig:diversity}~(b).

\setlength{\tabcolsep}{8pt}
\begin{table}[t]
  \caption{
    Quantitative comparison with GANSpace~\cite{harkonen2020ganspace}.
  }
  \vspace{2pt}
  \label{tab:comparison-ganspace}
  \centering\small
  \begin{tabular}{lccc}
    \toprule
                                         &           FID &    Re-scoring &    User Study \\ \midrule
    GANSpace~\cite{harkonen2020ganspace} &          7.43 &          0.33 &          41\% \\
    SeFa (Ours)                          & \textbf{7.36} & \textbf{0.38} & \textbf{59\%} \\ \bottomrule
  \end{tabular}
  \vspace{-10pt}
\end{table}

\subsection{Comparison with Unsupervised Baselines}\label{subsec:unsupervise-comparison}
We compare our method with some unsupervised alternatives, including the sampling-based method~\cite{harkonen2020ganspace} and the learning-based method~\cite{infogan}.
The major difference is that SeFa works as a closed-form solution, which is independent of any kind of data sampling or model training.

\vspace{2pt}
\noindent\textbf{Comparison with Sampling-based Baseline.}
GANSpace \cite{harkonen2020ganspace} proposes to perform PCA on a collection of sampled data to find principal directions in the latent space.
In this part, we compare SeFa with GANSpace on the StyleGAN model trained on FF-HQ dataset~\cite{stylegan}.
Fig.~\ref{fig:comparison-ganspace} visualizes some qualitative comparison results, where the semantics found by SeFa lead to a more precise control.
For example, when changing face pose, SeFa better preserves the identity and skin color.
We also quantitatively compare these two approaches with FID~\cite{fid}, re-scoring analysis, and user study.
Here, users are asked which approach changes a particular attribute more adequately on $2K$ manipulations.
Results are shown in Tab.~\ref{tab:comparison-ganspace}.
SeFa and GANSpace show close FID score since this is mostly determined by the generator itself as well as the manipulation model in Eq.~\eqref{eq:manipulation}, which are shared by these two methods.
But SeFa outperforms GANSpace on attribute re-scoring and user study.

\vspace{2pt}
\noindent\textbf{Comparison with Learning-based Baseline.}
InfoGAN~\cite{infogan} proposed to explicitly learn a factorized representation by introducing a regularizer to maximize the mutual information between the output image and the input latent code.
We compare our method with the Info-PGGAN model~\cite{progressive_infogan}, which trains the native PGGAN~\cite{pggan} with the information regularizer~\cite{infogan}.
Fig.~\ref{fig:comparison-infogan} shows the comparison results.
We can tell that the semantics identified by SeFa through a closed-form factorization on pre-trained weights are more accurate than those learned from Info-PGGAN.
Taking pose manipulation as an example, the hair color varies when using Info-PGGAN for editing.
By contrast, SeFa achieves a more precise control.%
\footnote{We use different samples for Info-PGGAN~\cite{progressive_infogan} and SeFa because Info-PGGAN requires model retraining, leading to a different model from the one that is officially released by~\cite{pggan}. As a result, it is hard to produce the same face with these two different models.}

\begin{figure}[t]
  \centering
  \includegraphics[width=1.0\linewidth]{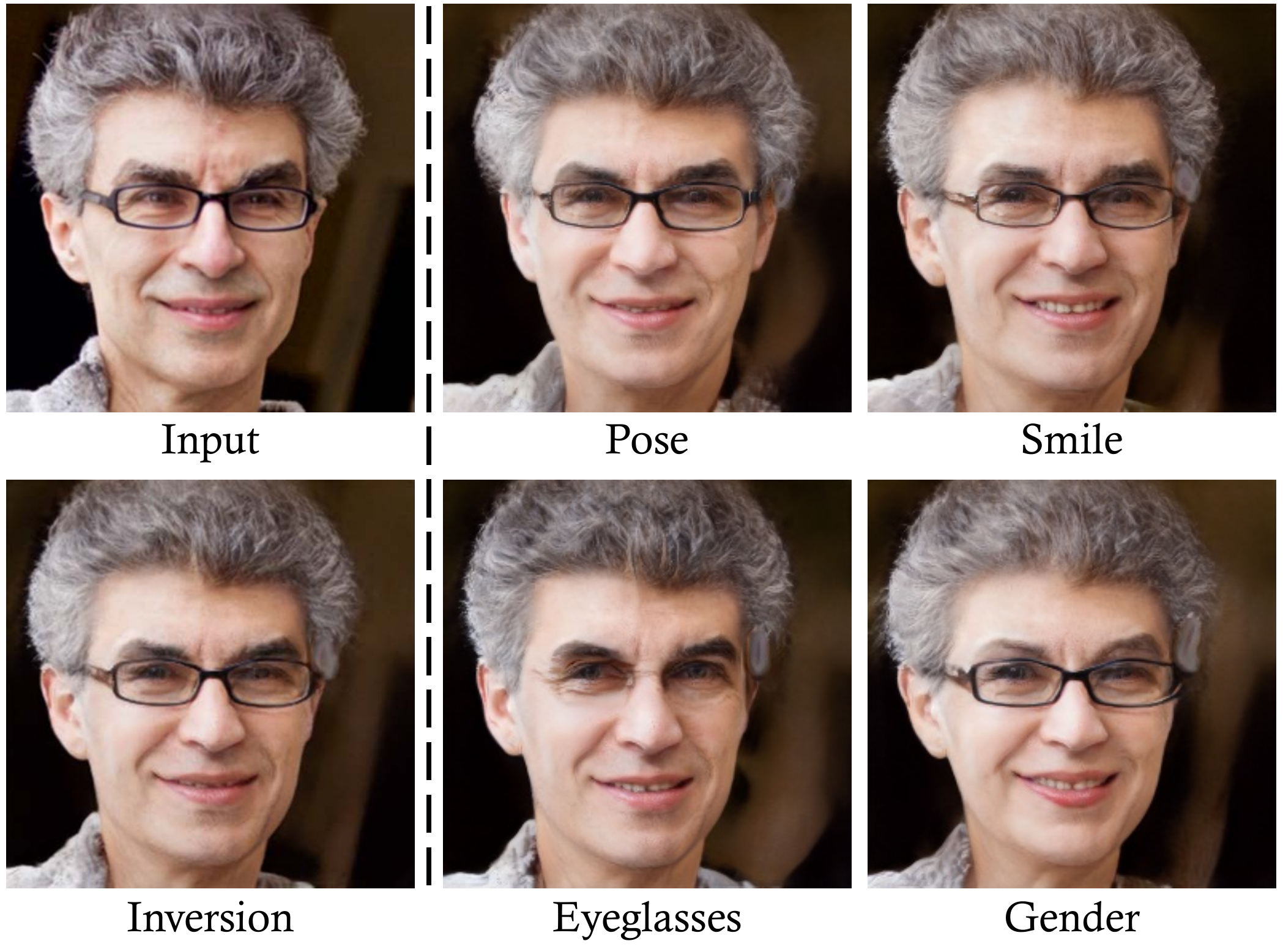}
  \caption{
    \textbf{Real image editing} with respect to various facial attributes.
    All semantics are found with the proposed SeFa.
    GAN inversion~\cite{zhu2020indomain} is used to project the target real image back to the latent space of StyleGAN~\cite{stylegan}.
  }
  \label{fig:real}
  \vspace{-5pt}
\end{figure}

\subsection{Real Image Editing}\label{subsec:real}
In this part, we verify that the latent semantics revealed by SeFa is applicable for real image editing.
Since the generator lacks the inference ability to take a real image as the input, we involve GAN inversion~\cite{gu2020image,zhu2020indomain} approaches into our algorithm.
More concretely, given a target image to edit, we first project it back to the latent space, and then use the variation factor found by SeFa to modulate the inverted code.
Fig.~\ref{fig:real} shows some examples, where SeFa shows satisfying performance.
For example, we manage to remove eyeglasses from the input images and also alter the face pose.
It suggests that SeFa is capable of discovering the directions of the latent space which are generalizable for real image editing.

\section{Conclusion}\label{sec:conclusion}
In this work we propose a closed-form solution to factorizing the latent semantics learned by GANs.
Extensive experiments demonstrate the great power of our algorithm in identifying versatile semantics from different types of GAN models in an unsupervised manner.

\vspace{2pt}
\noindent\textbf{Acknowledgements:}
This work is supported in part by the Early Career Scheme (ECS) through the Research Grants Council (RGC) of Hong Kong under Grant No.24206219, CUHK FoE RSFS Grant, SenseTime Collaborative Grant and Centre for Perceptual and Interactive Intelligence (CPII) Ltd under the Innovation and Technology Fund.

{\small
\bibliographystyle{ieee_fullname}
\bibliography{ref}
}

\end{document}